\documentclass[10pt,twocolumn,letterpaper]{article}

\usepackage[pagenumbers]{wacv} %

\usepackage{graphicx}
\usepackage{amsmath}
\usepackage{amssymb}
\usepackage{booktabs}

\usepackage{amsmath,amsfonts}
\usepackage{algorithmic}
\usepackage{algorithm}
\usepackage{array}
\usepackage{textcomp}
\usepackage{stfloats}
\usepackage{url}
\usepackage{verbatim}
\usepackage{graphicx}
\usepackage{cite}
\usepackage{multirow}
\usepackage{booktabs}

\usepackage[accsupp]{axessibility}  %

\usepackage[pagebackref,breaklinks,colorlinks]{hyperref}

\usepackage[capitalize]{cleveref}
\crefname{section}{Sec.}{Secs.}
\Crefname{section}{Section}{Sections}
\Crefname{table}{Table}{Tables}
\crefname{table}{Tab.}{Tabs.}

\def\wacvPaperID{xxxx} %
\def\confName{WACV}
\def\confYear{2024}

\newcommand\scalemath[2]{\scalebox{#1}{\mbox{\ensuremath{\displaystyle #2}}}}

\begin{document}

\setlength{\abovedisplayskip}{-6pt}
\setlength{\belowdisplayskip}{-2pt}
\setlength{\abovedisplayshortskip}{0pt}
\setlength{\belowdisplayshortskip}{-2pt}

\title{Semi-Supervised Semantic Depth Estimation using Symbiotic Transformer and NearFarMix Augmentation}

\author{Md Awsafur Rahman, Shaikh Anowarul Fattah\\
Dept. of EEE, BUET, Bangladesh\\
{\tt\small \{1706066,fattah\}@eee.buet.ac.bd}
}
\maketitle

\begin{abstract}

In computer vision, depth estimation is crucial for domains like robotics, autonomous vehicles, augmented reality, and virtual reality. Integrating semantics with depth enhances scene understanding through reciprocal information sharing. However, the scarcity of semantic information in datasets poses challenges. Existing convolutional approaches with limited local receptive fields hinder the full utilization of the symbiotic potential between depth and semantics. This paper introduces a dataset-invariant semi-supervised strategy to address the scarcity of semantic information. It proposes the Depth Semantics Symbiosis module, leveraging the Symbiotic Transformer for achieving comprehensive mutual awareness by information exchange within both local and global contexts. Additionally, a novel augmentation, NearFarMix is introduced to combat overfitting and compensate both depth-semantic tasks by strategically merging regions from two images, generating diverse and structurally consistent samples with enhanced control. Extensive experiments on NYU-Depth-V2 and KITTI datasets demonstrate the superiority of our proposed techniques in indoor and outdoor environments.

\end{abstract}

\section{Introduction}

The integration of depth information, crucial for 3D scene understanding, significantly contributes to applications such as bokeh, autonomous vehicles, robotics, and augmented-virtual reality~\cite{bokeh, car, robot, ar}. Combined with semantic information, it enables comprehensive scene interpretation and fosters a symbiotic relationship~\cite{multitask01, multitask02, multitask03}. However, obtaining this data from sensors presents formidable challenges, including high costs, sparse data, and inflexibility. As a result, there is growing interest in multi-task systems for interpreting depth and semantics from a single image. Monocular depth estimation remains an ill-posed problem~\cite{illpose01}, diluting 3D perception in 2D images. The scarcity of semantic labels in datasets further complicates multi-tasking. Thus, efficient solutions are a pivotal research domain.

Traditional methods for acquiring depth and semantic information have employed Self-Supervised Depth Estimation (SSDE) and Semi-supervised Semantic Segmentation (Semi-Seg)~\cite{multitask03, AffineMix, DepthMix, SW-Map, SIG, AG-MMD} to reduce reliance on labeled data. However, SSDE requires supplementary data like stereo imagery and motion artifacts, limiting its applicability. In terms of performance, SSDE lags behind supervised learning alternatives~\cite{iDisc, PixelFormer, AdaBin, NewCRFs, DPT, TransDepth}. Paradoxically, state-of-the-art supervised approaches neglect the symbiotic relationship between depth and semantics. Current methodologies resort to CNN-based strategies~\cite{AffineMix, DepthMix, SW-Map, SIG, AG-MMD}, which suffer from limited global context awareness~\cite{tradeoff}, leading to localized symbiosis and unresolved issues like blurred object boundaries and inadequate object-background contrast.

Augmentation has been employed to combat overfitting~\cite{ClassMix, CutMix}. However, applying it to depth estimation faces challenges, as conventional augmentations risk invalidating depth maps due to object scale and orientation alteration. Semantics-specific augmentations involving copy-pasting consider depth but compromise its integrity post-blending. Depth-specific augmentations aim to enhance depth but impair semantics due to discontinuity. In a depth-semantics multi-task system, a dual-favoring augmentation approach remains elusive.

This paper presents a novel approach to address the aforementioned issues, namely the absence of semantic labels, ineffective symbiosis, and lack of a dual-compensating depth-semantics augmentation. The key contributions of the proposed methodology can be distilled into the following points:

\begin{enumerate}
\item The introduced semi-supervised strategy not only tackles the prevalent issue of semantic label scarcity across datasets but also guarantees a dataset-agnostic architecture thus improving model's practicality.

\item The proposed Symbiotic Transformer effectively exploits the symbiotic relationship between depth and semantics, facilitating information exchange across both local and global contexts, thereby fostering comprehensive mutual awareness. 

\item A novel augmentation namely NearFarMix is proposed to address the critical issue of overfitting in depth estimation, while also improving semantic tasks. This technique tackles challenges common to both depth and semantics, including compromised integrity, limited diversity, and inconsistencies-discontinuities in semantics while promoting the generation of controlled samples.
\end{enumerate}

\section{Related Work}
\label{sec:related-work}

\textbf{Supervised Depth Estimation}: Eigen et al.~\cite{eigen} pioneered a coarse-to-fine approach, followed by Laina et al.~\cite{laina} with a fully CNN-based residual network. Transformer-based supervised methods~\cite{TransDepth, SwinDepth, AdaBin, PixelFormer, NewCRFs} have dominated due to their ability to capture long-range dependencies. iDisc~\cite{iDisc} utilizes internal discretized representations for improved performance. Notably, state-of-the-art methods neglect the symbiosis between depth and semantics.

\textbf{Semi-supervised Semantic Segmentation}: Semi-supervised Learning~\cite{noisy-student} is employed in segmentation tasks. Approaches like AG-MMD, SIG, AffineMix, DepthMix~\cite{AG-MMD, SIG, AffineMix, DepthMix} use a pre-trained teacher model to obtain semantic pseudo labels. However, relying on convolutional teacher models limits the global semantic interpretation.

\textbf{Depth-Semantics Symbiosis}: Current methodologies~\cite{AffineMix, DepthMix, SIG, depthmask-domain1} explore depth-semantics symbiosis with self-supervised depth estimation and convolutional information sharing. SW-Map~\cite{SW-Map} utilizes dual encoder-decoder systems, SIG~\cite{SIG} incorporates decoded semantic features into depth features, AG-MMD~\cite{AG-MMD} employs gated attention, and CCAM~\cite{AffineMix} uses SE-based channel-attention. Limited global context in spatial or channel dimensions constrains the potential of symbiosis.

\textbf{Augmentation for Depth-Semantics}: Semantics-centric augmentations like CutMix, ClassMix~\cite{CutMix, ClassMix} blend images, while DepthMix, AffineMix~\cite{DepthMix, AffineMix} integrate depth information. However, integrity compromise and structural disparities limit their suitability. Depth-oriented augmentations like Cut-Depth, GLPDepth~\cite{Cut-Depth, GLPDepth} enhance depth at the expense of semantic performance, introducing discontinuity and inconsistency.

In essence, self-supervised depth estimation pales compared to transformer-based supervised counterparts. Semi-supervised semantic segmentation falls short, tethered by only locally-aware teacher model. Current depth-semantics symbiosis methods inadequately exploit the symbiotic potential due to a lack of global context due to their convolutional short-sighted nature. Finally, depth-semantics augmentations often sacrifice depth or semantics, highlighting a struggle to uphold object integrity and structural coherence.

\begin{figure*}[t]
    \centering
\captionsetup{justification=justified,singlelinecheck=false}
    \includegraphics[scale=0.072]{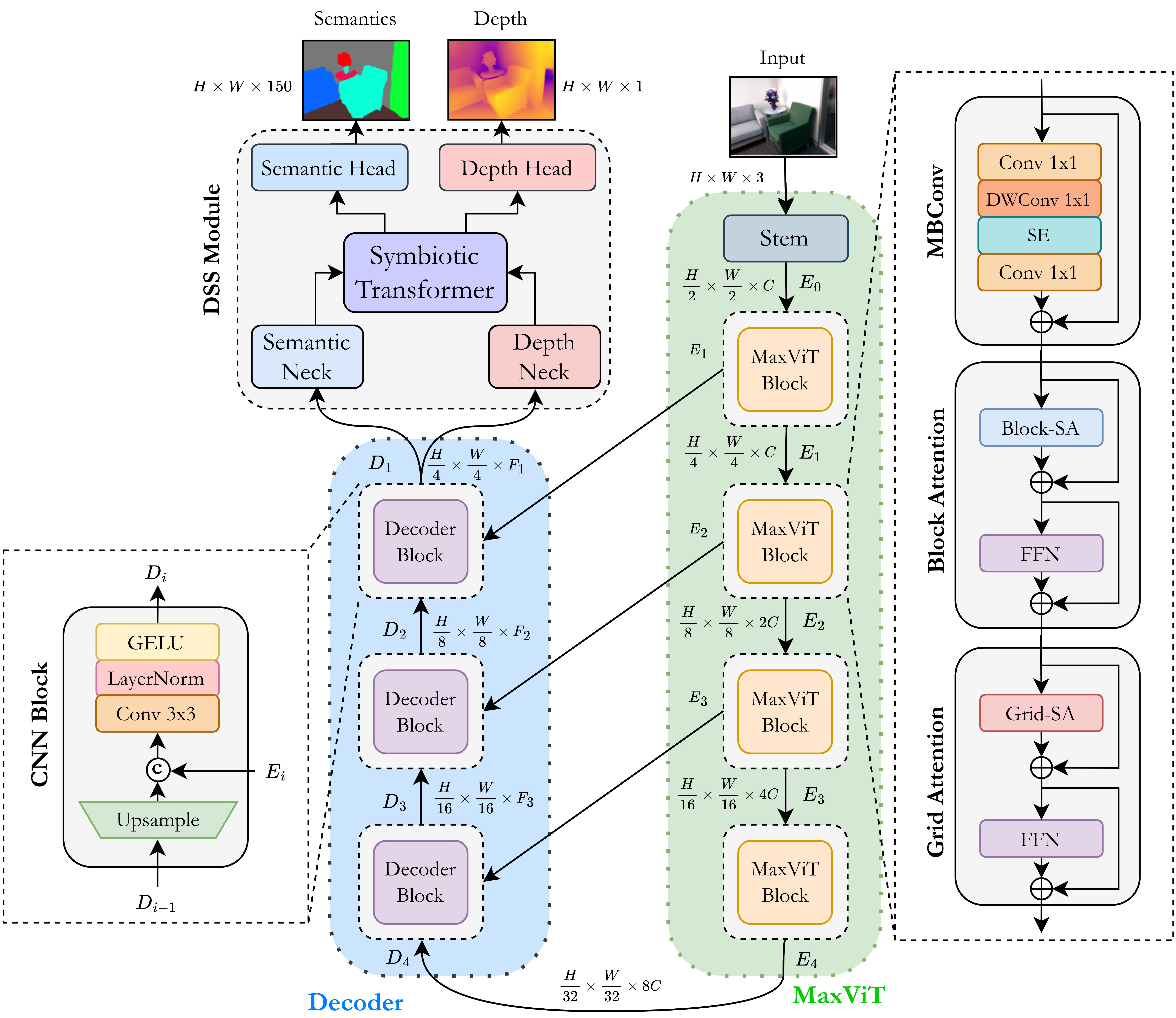}
    \caption{Proposed architecture overview. Input image is processed by Max-ViT encoder and CNN decoder. Then, resultant features undergo the Depth Semantics Symbiosis (DSS) module for joint Depth and Semantics generation via local-global information sharing.}
    \label{fig:arch}
\end{figure*}

\section{Methodology}
\label{sec:methodology}

The proposed research presents a unified framework for jointly modeling the depth map and generating a semantic mask of an RGB image. Given an input image $I \in \mathbb{R}^{H \times W \times 3}$, the developed model $\mathcal{F}(\Theta)$ predicts probability maps: $\hat{p}_d \in [0, 1]^{H \times W \times 1}$ for maximum depth likelihood and $\hat{p}_s \in [0, 1]^{H \times W \times 150}$ for 150 distinct classes in the semantic mask. The depth map is obtained by, $\hat{D} = \hat{p}_d \odot \textit{max\_depth}$. Finally, the semantic mask is obtained by, $\hat{S} = \operatorname{Max\_Index}(\hat{p}_s, \text{axis}=2)$.

\subsection{Proposed Semi-supervised Semantic Depth Estimation}

\begin{figure}[t]
    \centering
    \includegraphics[scale=0.07]{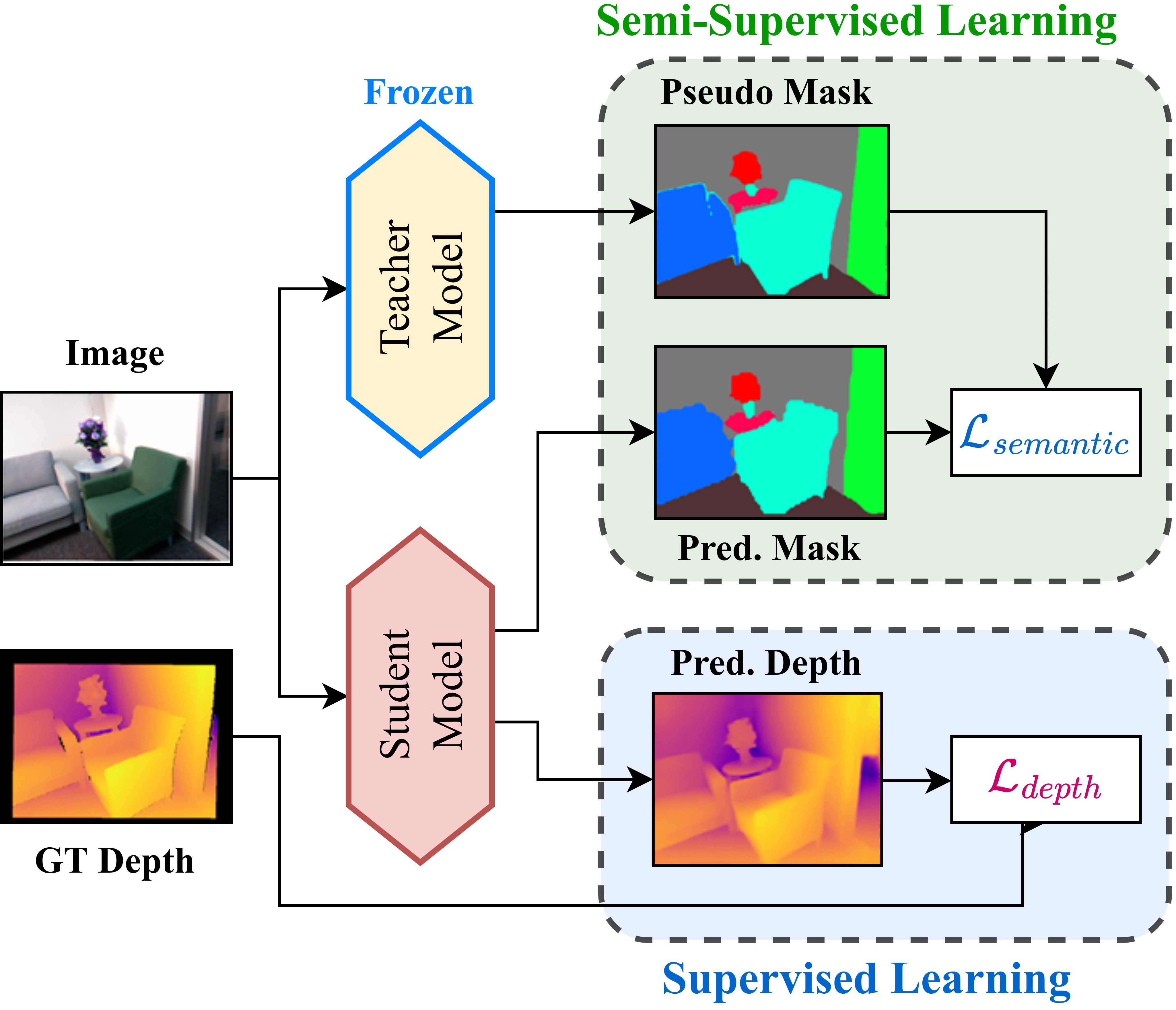}
\captionsetup{justification=raggedright,singlelinecheck=false}
    \caption{Proposed semi-supervised strategy. The Student model generates supervised Depth via ground-truth and semi-supervised Semantics via Teacher model.}
    \label{fig:semisup}
    \vspace*{-0.4cm}
\end{figure}

The proposed method (Figure \ref{fig:semisup}) combines supervised and semi-supervised learning to generate depth maps and semantic masks simultaneously. The semi-supervised component employs OneFormer~\cite{OneFormer}, a Transformer-based model pre-trained on ADE20K~\cite{ADE20K}, as the teacher model. It produces a pseudo mask utilized by the student model to capture semantics, fostering symbiosis. Simultaneously, the student model learns depth information in a supervised manner using ground truth data. The architecture adapts to the number of classes, but the teacher model maintains a fixed number of classes across datasets, ensuring a dataset-invariant architecture for the student model. The semi-supervised nature enables its application to datasets with only depth information~\cite{kitti, Make3D, Sun3D, Diode}, without requiring semantic annotations.

\subsection{Architecture Overview}

The proposed scheme employs a shared encoder-decoder structure enhanced by a Depth Semantics Symbiosis (DSS) module (\cref{subsec:dms}) for concurrent depth-semantic map generation. The ImageNet-pretrained Max-ViT \cite{MaxViT} encoder, integrated with Block and Grid attention mechanisms, captures spatial dynamics. Leveraging skip connections, the decoder refines multilevel encoder outputs ($\mathbf{E_i}$, $i \in [0, 4]$) prior to DSS application. Each decoder level involves upscaling the prior layer ($\mathbf{D_{i-1}}$), merging it with the corresponding encoder output ($\mathbf{E_{i}}$), and undergoing convolution, normalization, and GELU activation to yield $\mathbf{D_i}$, given by:

\begin{equation}
\vspace*{-0.05cm}
\begin{aligned}
\label{eq:decoder}
\mathbf{D'_{i-1}} &= \operatorname{Concat}(\mathbf{E_{i}}, \operatorname{Upsample}(\mathbf{D_{i-1}})) \\
\mathbf{D_{i}} &= \operatorname{GELU}(\operatorname{LayerNorm}(\operatorname{Conv_{3\times3}}(\mathbf{D'_{i-1}})))\\
\end{aligned}
\end{equation}

\subsection{Proposed Depth Semantics Symbiosis Module}
\label{subsec:dms}

\begin{figure}[h]
    \centering
    \captionsetup{justification=raggedright}
    \includegraphics[scale=0.073]{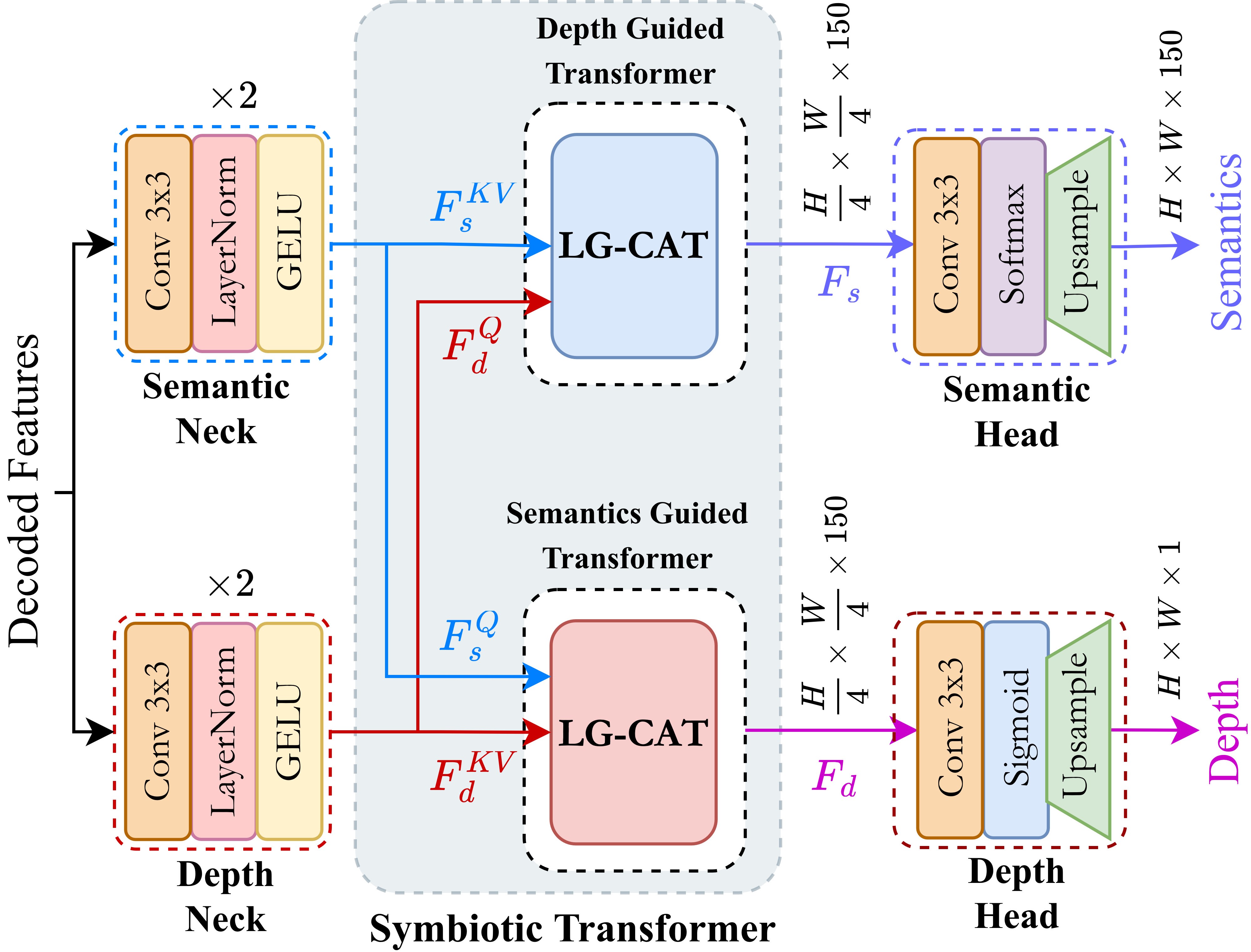}
    \caption{Proposed Depth Semantics Symbiosis (DSS) Module. The decoded features pass through Depth / Semantics Necks, Symbiotic Transformer for local-global sharing, producing Depth and Semantics with separate Heads.}
    \label{fig:dms}
\end{figure}

The proposed Depth Semantics Symbiosis Module (DSS) leverages the Symbiotic Transformer (refer to Section \ref{subsubsec:symbiotic-transformer}) to exploit the symbiotic relationship between depth and semantic features. In Figure \ref{fig:dms}, shared decoded features undergo individual processing through distinct $\operatorname{Neck}$ blocks, generating independent depth and semantic features. These features are then fed into the Symbiotic Transformer for information exchange within local and global contexts, resulting in depth-aware-semantic and semantic-aware-depth features. Separate $\operatorname{Head}$ blocks refine these mutually-aware features, generating the depth map and semantic mask. The mathematical representation of the Depth Semantics Symbiosis Module (DSS) is encapsulated in the following equation:

\begin{equation}
\begin{aligned}
\label{eq:dms}
\mathbf{F_s^{KV}} &= \mathbf{F_s^{Q}}  = \operatorname{Semantic-Neck}(\mathbf{D_1})\\
\mathbf{F_d^{KV}} &= \mathbf{F_d^{Q}}  = \operatorname{Depth-Neck}(\mathbf{D_1})\\
\mathbf{F_s} &= \operatorname{LG-CAT}(\mathbf{F_d^{Q}}, \mathbf{F_s^{KV}}) \\
\mathbf{F_d} &= \operatorname{LG-CAT}(\mathbf{F_s^{Q}}, \mathbf{F_d^{KV}}) \\
\mathbf{Semantics} &= \operatorname{Semantic-Head}(\mathbf{F_s})\\
\mathbf{Depth} &= \operatorname{Depth-Head}(\mathbf{F_d})\\
\end{aligned}
\end{equation}
\vspace*{0.2cm}

where $\mathbf{D_1}$ is the final decoder output. $\mathbf{F_s^{Q}}$ and $\mathbf{F_s^{KV}}$ are semantics' query and key-value features, respectively. $\mathbf{F_d^{Q}}$ and $\mathbf{F_d^{KV}}$ represent depth's query and key-value features. Enhanced features $\mathbf{F_s}$ and $\mathbf{F_d}$ capture depth-aware semantic and semantics-aware depth information, respectively.

\begin{figure*}[h]
    \centering
    \captionsetup{justification=raggedright}
    \includegraphics[scale=0.082]{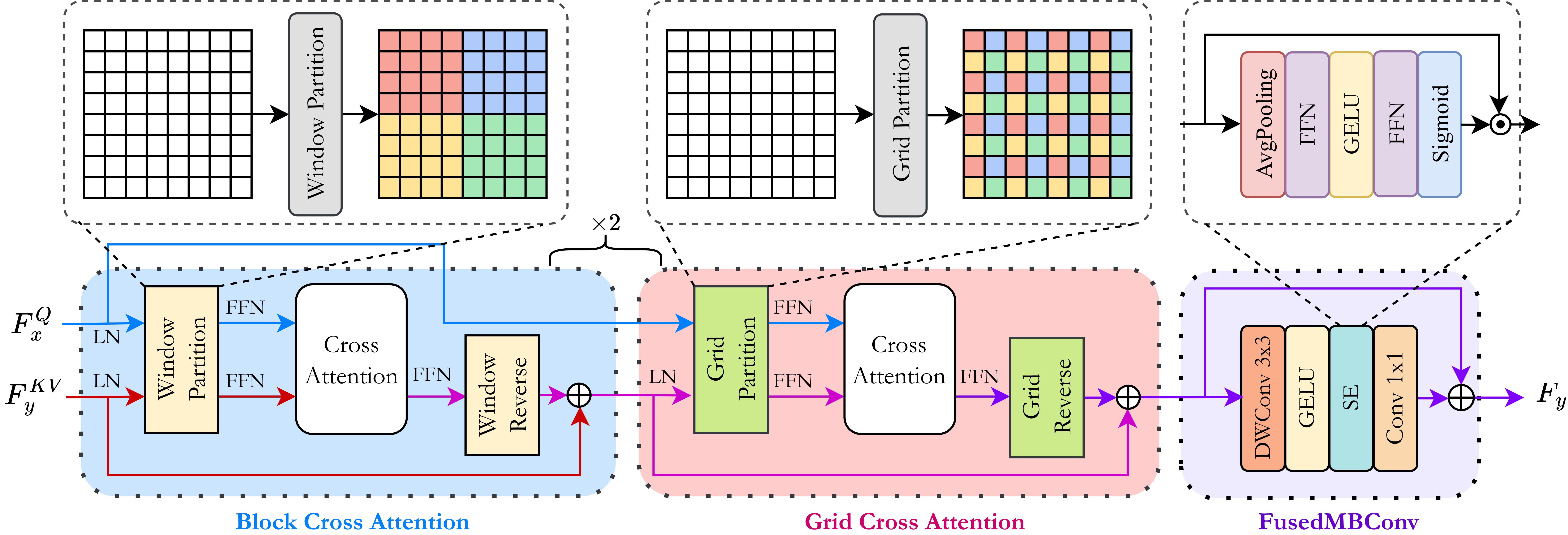}
    \caption{Proposed Local Global Cross Attention Transformer (LG-CAT). Input features undergo Block-Grid Cross Attentions for local-global spatial awareness and FusedMBConv block for channelwise awareness.}
    \label{fig:lg-cat}
\end{figure*}

\subsubsection{Depth-Semantics Neck}
The final decoded features ($\mathbf{D_1}$) are fed into specialized Neck blocks for depth and semantics. These Neck blocks, with distinct weights, generate independent depth and semantic features. Each Neck block includes a $\operatorname{Conv}_{3\times3}$ layer, $\operatorname{LayerNorm}$, and $\operatorname{GELU}$. Mathematically:

\begin{equation}
\operatorname{Neck}(\mathbf{x}) = \begin{bmatrix}
\mathbf{x} = \operatorname{Conv}_{3\times3}(\mathbf{x}) \\
\mathbf{x} = \operatorname{LayerNorm}(\mathbf{x}) \\
\mathbf{x} = \operatorname{GELU}(\mathbf{x}) \\
\end{bmatrix} \times 2
\end{equation}

\subsubsection{Symbiotic Transformer}
\label{subsubsec:symbiotic-transformer}

Figure \ref{fig:dms} provides an overview of the proposed module that utilizes depth and semantic features as inputs. It consists of two specialized transformers: Semantics Guided Transformer (SGT) and Depth Guided Transformer (DGT), both instances of Local-Global Cross Attention Transformers (LG-CAT). SGT treats semantic features as query features ($\mathbf{F_s^{Q}}$) and depth features as key-value features ($\mathbf{F_d^{KV}}$), while DGT operates in the opposite manner. These features are transformed into Queries ($\mathbf{Q_s}$, $\mathbf{Q_d}$) and Key-Values ($\mathbf{K_s}$, $\mathbf{K_d}$, $\mathbf{V_s}$, $\mathbf{V_d}$) for cross-attention. The cross-attention operation can be expressed as:

\begin{align}
\label{eq:cross-attention}
& \operatorname{CA}\left(\mathbf{Q_x}, \mathbf{K_y}, \mathbf{V_y}\right)=\mathcal{S}mx\left(\frac{\mathbf{Q_x} {\mathbf{K_y}}^T}{\sqrt{\mathbf{d}}}+\mathbf{B}\right)\mathbf{V_y}
\end{align}
\vspace*{0.1cm}

Here, $\mathbf{Q_x}$ represents query features, $\mathbf{K_y}$ and $\mathbf{V_y}$ represent key and value features, $\mathcal{S}mx$ denotes softmax, and $\mathbf{d}$ is the query/key dimension. $\mathbf{B}$ represents relative positional bias, sampled similar to~\cite{GCViT}.

The LG-CAT block (Figure \ref{fig:lg-cat}) comprises three modules: Block Cross Attention (BCA), Grid Cross Attention (GCA), and FusedMBConv (FMBC). BCA partitions query and key-value features into local dense windows, enabling local symbiotic enrichment. GCA partitions features into global sparse windows, facilitating global information sharing beyond limited receptive fields. FMBC enables inter-channel interactions while maintaining important properties.

The Symbiotic Transformer block can be mathematically expressed as:

\vspace*{0.1cm}
\begin{equation}
\scalemath{0.90}{
\begin{aligned}
\label{eq:symbiotic-transformer}
&\left[
\begin{aligned}
\mathbf{{}_{i}F_y^{KV}} & = \operatorname{Block-Cross-Attention}(\mathbf{F_x^{Q}}, \mathbf{{}_{i-1}F_y^{KV}})\\
\mathbf{{}_{i+1}F_y^{KV}} & = \operatorname{Grid-Cross-Attention}(\mathbf{F_x^{Q}}, \mathbf{{}_{i}F_y^{KV}}) \\
\end{aligned}
\right]_{\times N_s} \\
& \quad \quad \quad \mathbf{F_y} = \operatorname{FusedMBConv}(\mathbf{{}_{N+1}F_y^{KV}})
\end{aligned}}
\end{equation}
\vspace*{0.05cm}

Here, $i$ denotes the iteration, ranging from $2$ to $N_s$. $\mathbf{F_x^{Q}}$ represents query features of $\mathbf{x}$, $\mathbf{F_y^{KV}}$ denotes key-value features of $\mathbf{y}$, and $\mathbf{F_y}$ signifies the output features contextualized by $\mathbf{x}$. Specifically, for SGT, $\mathbf{x}=\mathbf{s}$ and $\mathbf{y}=\mathbf{d}$, while for DGT, $\mathbf{x}=\mathbf{d}$ and $\mathbf{y}=\mathbf{s}$.

\subsubsection{Depth-Semantics Head}

Enhanced depth and semantic features ($\mathbf{F_d}, \mathbf{F_s}$) from the DSS module pass through separate Depth and Semantics Head blocks. The Depth Head applies convolution ($\operatorname{Conv}_{3\times3}$), sigmoid activation, and $4\times$ upsampling, yielding the final depth map ($\left(H, W, 1\right)$). The Semantics Head performs convolution, softmax activation, and $4\times$ upsampling, generating the semantic mask ($\left(H, W, 150\right)$). The Head blocks can be represented as:

\begin{equation}
\operatorname{Head}(\mathbf{x}) = \begin{bmatrix}
\mathbf{x} = \operatorname{Conv}_{3\times3}(\mathbf{x}) \\
\mathbf{x} = \operatorname{Activation}(\mathbf{x}) \\
\mathbf{x} = \operatorname{Upsample}_{4 \times 4}(\mathbf{x}) \\
\end{bmatrix} \times 1
\end{equation}

\subsection{Proposed NearFarMix Augmentation}

The NearFarMix algorithm operates on input elements $I_1$ and $I_2$, each comprising an image, a semantic mask, and depth information: $I_i = \begin{bmatrix} \text{Image}_i, \text{Mask}_i, \text{Depth}_i \end{bmatrix}$, where $i$ denotes the input identifier. $I_1$ labels the $Near$ region below a depth threshold $thr$, while $I_2$ labels the pre-$Far$ region exceeding the threshold. The algorithm combines the Near and pre-Far regions additively, eliminates the overlapping Far region, and fills the mutually exclusive region from $I_2$. Ultimately, the pre-Far, overlap, and exclusive regions contribute to only Far regions. The augmented output is represented as $I' = [\text{Image}', \text{Mask}', \text{Depth}']$. Mathematically, this can be expressed as:

\begin{equation}
\scalemath{0.90}{
\begin{aligned}
\label{eq:near-far-mix}
M_1 &= (Depth_1 \leq thr); \quad M_2 = (Depth_2 > thr) \\
M_o &= - (M_1 \odot M_2); \quad M_e = (1 - M_1) \odot (1 - M_2) \\
I' &= (I_1 \odot M_1) +  (I_2 \odot M_2) + (I_2 \odot M_o) + (I_2 \odot M_e)
\end{aligned}}
\end{equation}
\vspace*{0.1cm}

where $\odot$ denotes element-wise multiplication, $M_1$ and $M_2$ denote binary masks derived from the depth maps and the depth threshold $thr$. $M_o$ represents the overlapped region, while $M_e$ denotes the mutually exclusive region. Finally, the augmented output $I'$ is obtained by summing the four masked components. More details in supplementary (suppl.) material.

In figures \ref{fig:nearfarmix-kitti-quality}, and \ref{fig:nearfarmix-nyuv2-quality}, augmented samples is presented generated by proposed NearFarMix. The resultant $Image'$, a synthesis of $Image_1$ and $Image_2$, exemplifies both structural and semantic consistency while upholding object integrity. Existing semantics-centric methods~\cite{DepthMix, AffineMix} indiscriminately compare pixel depth values across two images, thereby risking object integrity due to their disregard for strategic blending and occlusion considerations. Conversely, depth-centric techniques \cite{Cut-Depth, GLPDepth} often induce semantic discord during the image-depth fusion.  In contrast, proposed NearFarMix augmentation strategically blends specific regions from two images, duly considering their depth values. This approach ensures both object integrity within the selected regions and overall semantic coherence. Governed by the depth threshold parameter, $thr$, NearFarMix presents flexible blending options and precise control over blend proportion, thereby elevating the diversity of the output.

\begin{figure*}[t]
    \centering
    \captionsetup{justification=centering}
    \includegraphics[scale=0.48]{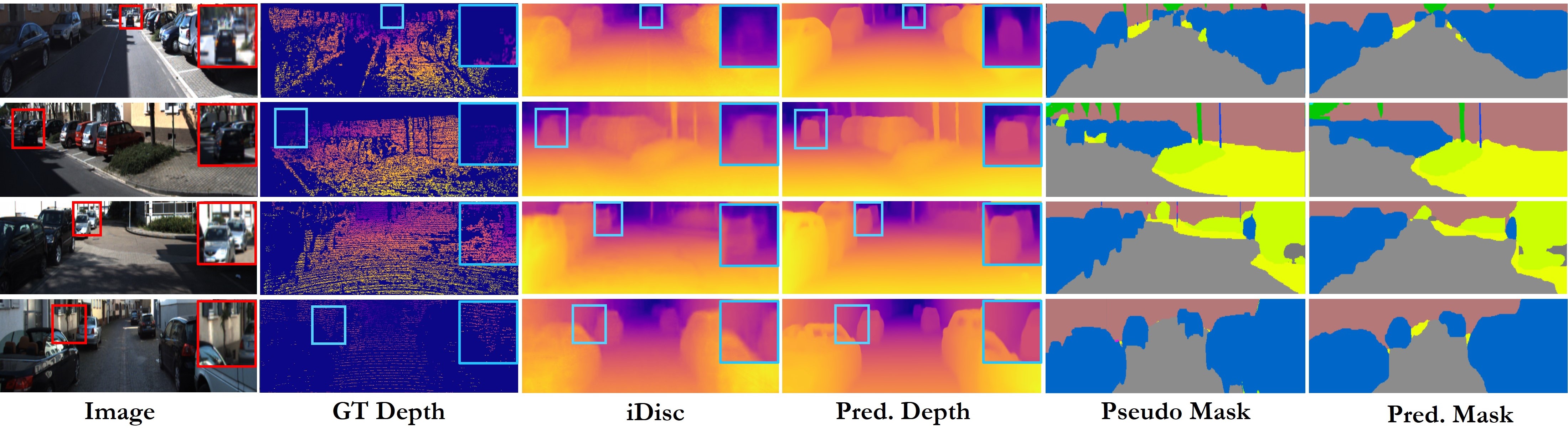}
    \caption{Qualitative results comparing proposed method to previous SOTA on KITTI dataset.}
    \label{fig:kitti-quality}
\end{figure*}

\begin{figure*}[h]
    \centering
    \captionsetup{justification=raggedright}
    \includegraphics[scale=0.57]{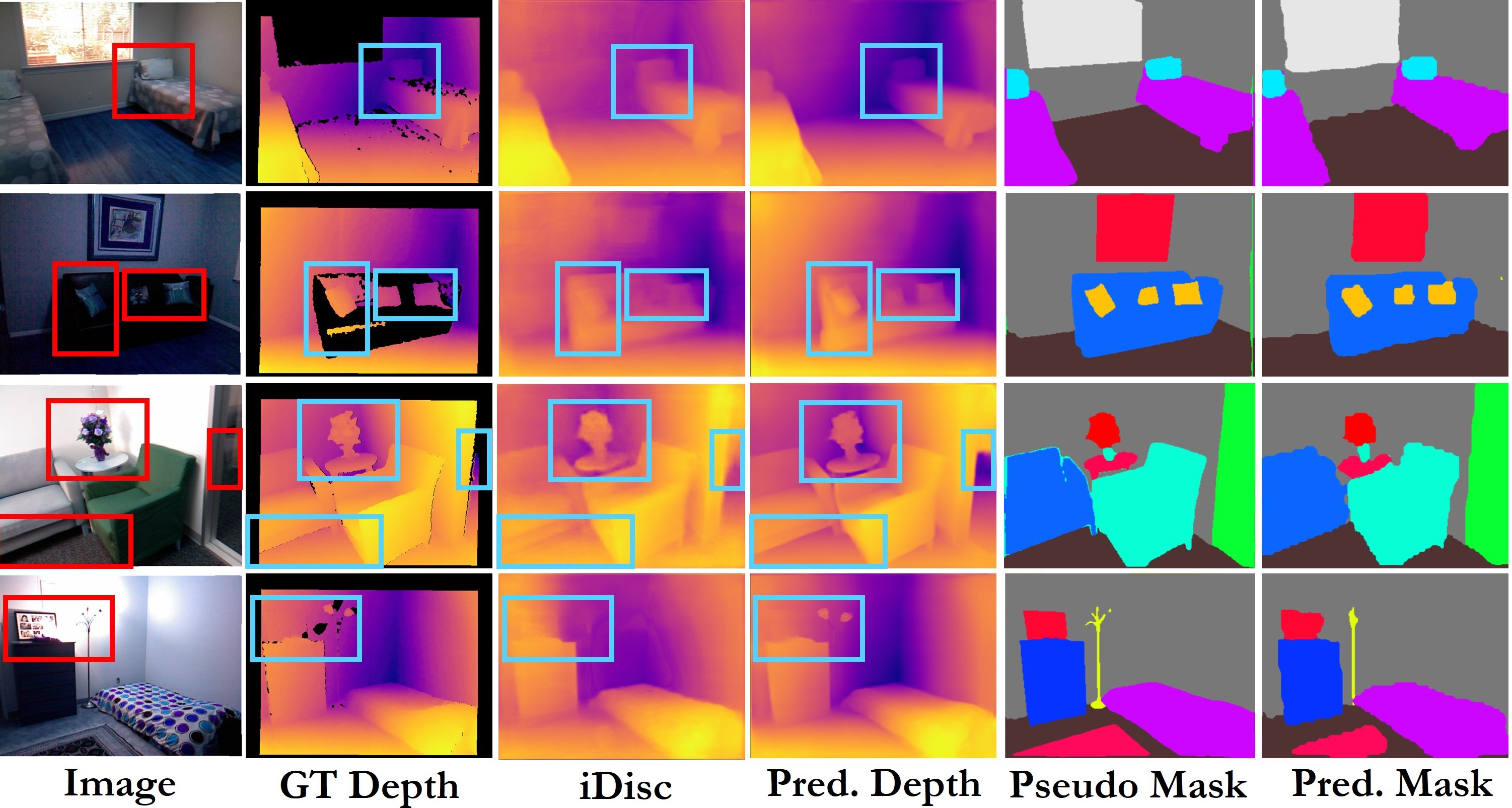}
    \caption{Qualitative results comparing proposed method to previous SOTA on NYUv2 dataset.}
    \label{fig:nyuv2-quality}
\end{figure*}

\subsection{Loss Function}

The multi-task architecture proposed uses the Scale-Invariant (SI) loss~\cite{eigen} and Jaccard (IoU) loss~\cite{jaccard} for depth estimation and semantic segmentation, respectively. Equation (\ref{eq:si-loss}) represents the SI loss, with $d_i$ as the ground truth depth, $\hat{d}_i$ as the estimated depth, $T$ as the count of valid ground truth pixels, and $g_i=\log_e(\hat{d}_i)-\log_e(d_i)$. Parameters $\lambda$ and $\alpha$ are 0.85 and 10.

\begin{equation}
\label{eq:si-loss}
\mathcal{L}_{\text {depth}}=\alpha \sqrt{\frac{1}{T} \sum_{i} g_i^2-\frac{\lambda}{T^2}\left(\sum_i g_i\right)^2}
\end{equation}

The IoU loss, used for semantic segmentation, is presented in Equation (\ref{eq:jaccard-loss}), with $C$ representing classes (150 in total), $N$ denoting pixel count, and $m_{ij}$ and $\hat{m}_{ij}$ as the ground truth and predicted semantic labels for pixel $j$ and class $i$.

\begin{equation}
\scalemath{0.95}{
\begin{aligned}
\label{eq:jaccard-loss}
& \mathcal{L}_{\text {semantic}} = \frac{1}{C} \sum_{i=1}^{C} \left( 1 - \mathcal{L}_{iou}^i \right)\\
\mathcal{L}_{iou}^i &= \frac{\sum_{j=1}^N m_{ij} \cdot \hat{m}_{ij}}{\sum_{j=1}^N m_{ij} + \sum_{j=1}^N \hat{m}_{ij} - \sum_{j=1}^N m_{ij} \cdot \hat{m}_{ij} }
\end{aligned}}
\end{equation}
\vspace*{0.1cm}

The overall loss, amalgamating depth and semantic losses, is as follows:

\begin{equation}
\label{eq:total-loss}
\mathcal{L}_{\text {total}} = \mathcal{L}_{\text {depth}} + \mathcal{L}_{\text {semantic}}
\end{equation}

\section{Experiments}
\label{sec:experiments}

\subsection{Datasets}
The ADE20K~\cite{ADE20K} pre-training dataset contains 20K/2K (training/validation) images with 150 classes. The NYUv2 dataset~\cite{nyuv2}, comprised of 120K RGB-depth pairs from 464 unique indoor scenes, provides a valuable asset for the study. The methodology benefits from a prescribed split of 50K training images and 654 test images, conforms to a depth limit of 10 meters, and is trained at a resolution of $480 \times 640$. In contrast, the KITTI dataset~\cite{kitti}, based on Eigen et al.~\cite{eigen}, features stereo images from 61 distinct outdoor scenes, outlines a 26K/697 training/testing split, and extends the depth threshold to 80 meters. Training for this methodology occurs at a resolution of $704 \times 352$. 

\subsection{Implementation Details}

\textbf{Evaluation Metrics}: The proposed method evaluates depth estimation using metrics such as Average Relative error (Abs Rel), Root Mean Squared error (RMS), Average Logarithmic error ($\log _{10}$), Root Mean Squared Log error ($\text{RMS}_{log}$), and Threshold Accuracy ($\delta_i$)~\cite{eigen}. For semantic segmentation, the mean Intersection over Union (mIoU) is employed~\cite{jaccard}, quantifying the overlap between predicted and ground truth semantic maps. A threshold of 0.5 is applied to obtain a binary representation before computing the mIoU score.

\textbf{Network Architecture}: The proposed method adopts an ImageNet pre-trained Max-ViT-Large~\cite{MaxViT} encoder with channels per level $C = 128$. Decoder output channels are $F_i = 2^{i+6}$, where $i \in [1, 3]$. The depth of the Symbiotic Transformer, regulated by $N_s$, signifies Grid-Block attention repetitions; $N_s = 2$ is found empirically. For Block-Grid attentions,  $h=4$ heads, and $w=7$ window size is sued, with remaining parameters analogous to Max-ViT. More details in supplementary (suppl.) material.

\textbf{Training}: The encoder-decoder, excluding the Symbiotic Transformer module, is pre-trained on ADE20K~\cite{ADE20K} for 30 epochs to establish a favorable starting point and to prepare the encoder for feeding the decoder. The AdamW optimizer~\cite{AdamW} is applied with $\beta_1 = 0.9$, $\beta_2 = 0.999$, and weight decay follows the scheduler. For scheduler, learning rate commenced at $4 \times 10^{-6}$, linearly increased to $3 \times 10^{-5}$, and then reduced with a Cosine function. The Symbiosis Training procedure is executed over 20 epochs using 8 NVIDIA V100 GPUs and a batch size of 4. Augmentations including random horizontal flipping, grayscale, and jitter are applied, complemented by the proposed NearFarMix. During NearFarMix, threshold, $thr$, is sampled from uniform distribution, $\mathcal{U}\left(max(d_{\text{min}}, D_{\text{min}}), min(d_{\text{max}}, D_{\text{max}})\right)$. Here, $d_{\text{min}}, d_{\text{max}}$ refer to the image depth and $D_{\text{min}}, D_{\text{max}}$ to dataset-specific depths (e.g., 20m and 60m for KITTI, and 1.5m and 6.5m for NYUv2).

\subsection{Comparison with Existing Methods}

\subsubsection{Quantitative Comparison}
Tables \ref{tab:nyuv2-method} and \ref{tab:kitti-method} quantitatively compare the proposed method against existing techniques on the KITTI outdoor and NYUv2 indoor datasets, respectively. The data substantiate the superior performance of proposed method, consistently outperforming competitors across all metrics due to its effective utilization of depth-semantics symbiosis. Notably, while many extant methods attempt to harness symbiotic potential through self-supervised depth estimation strategies, these generally underperform compared to supervised methods and are consequently excluded from the tables. Nevertheless, their symbiotic modules are evaluated against proposed method in the ablation study (\cref{subsubsec: dms-ablation}).

\begin{table}
\centering
\caption{Comparison on NYUv2 dataset with existing methods. The best result is indicated in \textbf{bold}, second best is \underline{underlined}, and symbols $\uparrow$ or $\downarrow$ denote higher/lower values are preferable}
\label{tab:nyuv2-method}
\scalebox{0.84}{
\begin{tabular}{lcccc} 
\toprule
Method        & Abs Rel $\downarrow$ & RMS $\downarrow$ & $\log_{10}$ $\downarrow$ & $\delta_1 \uparrow$  \\ 
\midrule
Eigen et al.~\cite{eigen} & 0.158                & 0.641            & -                      & 0.769                \\
DORN~\cite{DORN}          & 0.115                & 0.509            & 0.051                  & 0.828                \\
BTS~\cite{BTS}            & 0.110                & 0.392            & 0.047                  & 0.885                \\
TransDepth~\cite{TransDepth}   & 0.106                & 0.365            & 0.045                  & 0.900                \\
DPT~\cite{DPT}            & 0.110                & 0.367            & 0.045                  & 0.904                \\
Adabins~\cite{AdaBin}     & 0.103                & 0.364            & 0.044                  & 0.903                \\
P3Depth~\cite{P3depth}    & 0.104                & 0.356            & 0.043                  & 0.898                \\
NeWCRFs~\cite{NewCRFs}    & 0.095                & 0.334            & 0.041                  & 0.922                \\
PixelFormer~\cite{PixelFormer} & 0.090        & 0.322    & 0.039          & 0.929        \\
iDisc~\cite{iDisc}         & \underline{0.086}    & \underline{0.313}    & \underline{0.037}          & \underline{0.940}                \\ 
\midrule
\textbf{Proposed}         & $\mathbf{0.080}$     & $\mathbf{0.289}$ & $\mathbf{0.034}$       & $\mathbf{0.948}$     \\
\bottomrule
\end{tabular}}
\end{table}

\begin{table}
\centering
\captionsetup{justification=justified,singlelinecheck=false}
\caption{Comparison on KITTI dataset with existing methods. The best result is indicated in \textbf{bold}, second best is \underline{underlined}, and symbols $\uparrow$ or $\downarrow$ denote higher/lower values are preferable}
\label{tab:kitti-method}
\scalebox{0.85}{
\begin{tabular}{lcccc} 
\toprule
Method            & Abs Rel $\downarrow$ & RMS $\downarrow$    & $\text{RMS}_{log} \downarrow$ & $\delta_1 \uparrow$  \\ 
\midrule
Eigen et al.~\cite{eigen}     & $0.203$              & $6.307$             & $0.270$                     & $0.702$              \\
DORN.~\cite{DORN}            & $0.072$              & $2.727$             & $0.120$                     & $0.932$              \\
BTS~\cite{BTS}              & $0.059$              & $2.756$             & $0.090$                     & $0.956$              \\
TransDepth~\cite{TransDepth}       & $0.064$              & $2.755$             & $0.098$                     & $0.956$              \\
Adabins~\cite{AdaBin}          & $0.058$              & $2.360$             & $0.088$                     & $0.964$              \\
DPT~\cite{DPT}              & $0.060$              & $2.573$             & $0.088$                     & $0.959$              \\
NeWCRFs~\cite{NewCRFs}          & $0.052$              & $2.129$             & $0.079$                     & $0.974$              \\
PixelFormer~\cite{PixelFormer}      & $0.051$  & $2.081$ & $0.077$         & $0.976$  \\
iDisc             & $\underline{0.050}$              & $\underline{2.067}$             & $\underline{0.077}$                     & $\underline{0.977}$              \\ 
\midrule
\textbf{Proposed} & $\mathbf{0.048}$     & $\mathbf{1.984}$    & $\mathbf{0.075}$            & $\mathbf{0.979}$     \\
\bottomrule
\end{tabular}}
\end{table}

\subsubsection{Qualitative Comparison}

Figures~\ref{fig:kitti-quality}, and ~\ref{fig:nyuv2-quality}  visually contrasts the proposed method with existing techniques on the KITTI and NYUv2 datasets. The areas of exceptional performance by our method are spotlighted within red (image) and blue (depth) rectangles. This qualitative comparison is grounded on three criteria: 1) Edge sharpness, 2) Object-background contrast, and 3) Similarity between prediction and ground truth. As underscored by the marked rectangles, the proposed method evidently demonstrates superiority in these aspects.

\subsection{Ablation Study}

\begin{table}
\centering
\captionsetup{justification=raggedright,singlelinecheck=false}
\caption{Ablation analysis of NearFarMix on KITTI and NYUv2 datasets. $\uparrow$ signifies improvement, $\downarrow$ degradation. Best outcomes are in \textbf{bold}; (*) denotes the proposed technique.}
\label{tab:nearfarmix-ablation}
\scalebox{0.78}{
\begin{tabular}{clcccc} 
\toprule
\multicolumn{1}{l}{Dataset} & Method                    & RMS $\downarrow$ & Abs Rel $\downarrow$ & $\delta_1 \uparrow$ & mIoU $\uparrow$   \\ 
\midrule
                            & Baseline                  & $2.045$          & $0.051$              & $0.977$             & $0.663$           \\
\multirow{7}{*}{KITTI}      & CutMix~\cite{CutMix}      & $2.441$          & $0.059$              & $0.962$             & $0.671$           \\
                            & ClassMix~\cite{ClassMix}  & $2.302$          & $0.056$              & $0.967$             & $0.693$           \\
                            & AffineMix~\cite{AffineMix}& $2.102$          & $0.053$              & $0.969$             & $0.711$           \\
                            & DepthMix~\cite{DepthMix}  & $2.104$          & $0.054$              & $0.968$             & $0.708$           \\
                            & CutDepth~\cite{Cut-Depth}                  & $2.068$          & $0.050$              & $0.973$             & $0.640$           \\
                            & V-CutDepth~\cite{GLPDepth}                & $2.065$          & $0.050$              & $0.975$             & $0.645$           \\ 
\cmidrule{2-6}
                            & \textbf{NearFarMix}$^{*}$ & $\mathbf{1.984}$ & $\mathbf{0.048}$     & $\mathbf{0.979}$    & $\mathbf{0.731}$  \\ 
\midrule
\multirow{8}{*}{NYUv2}      & Baseline                  & $0.296$          & $0.082$              & $0.945$             & $0.568$           \\
                            & CutMix~\cite{CutMix}      & $0.331$          & $0.094$              & $0.927$             & $0.572$           \\
                            & ClassMix~\cite{ClassMix}  & $0.320$          & $0.090$              & $0.930$             & $0.585$           \\
                            & AffineMix~\cite{AffineMix}& $0.314$          & $0.086$              & $0.941$             & $0.603$           \\
                            & DepthMix~\cite{DepthMix}  & $0.305$          & $0.084$              & $0.943$             & $0.601$           \\
                            & CutDepth~\cite{Cut-Depth}                  & $0.294$          & $0.082$              & $0.946$             & $0.542$           \\
                            & V-CutDepth~\cite{GLPDepth}               & $0.290$          & $0.081$              & $0.946$             & $0.546$           \\ 
\cmidrule{2-6}
                            & \textbf{NearFarMix}$^{*}$ & $\mathbf{0.289}$ & $\mathbf{0.080}$     & $\mathbf{0.948}$    & $\mathbf{0.620}$  \\
\bottomrule
\end{tabular}}
\end{table}

\subsubsection{Depth Semantics Symbiosis Module}
\label{subsubsec: dms-ablation}

Table~\ref{tab:dss-ablation} compares the DSS approach and other symbiotic methodologies on KITTI and NYUv2 datasets using depth and semantic metrics, with NearFarMix augmentation included. The results underscore the need for symbiosis, as the baseline model underperforms. Approaches like SW-Map~\cite{SW-Map} and SIG~\cite{SIG} integrate semantics into depth at various scales, while others like AG-MMD~\cite{AG-MMD} and CCAM~\cite{AffineMix} employ attention mechanisms yet fall short in global context consideration. In contrast, the DSS method, leveraging its transformer, excels in information exchange across contexts.

\begin{table}
\centering
\captionsetup{justification=raggedright,singlelinecheck=false}
\caption{Ablation analysis of DSS on KITTI and NYUv2 datasets. $\uparrow$ signifies improvement, $\downarrow$ degradation. Best outcomes are in \textbf{bold}; (*) denotes the proposed technique}
\label{tab:dss-ablation}
\scalebox{0.80}{
\begin{tabular}{clcccc} 
\toprule
\multicolumn{1}{l}{Dataset} & \multicolumn{1}{c}{Method}                                                & RMS $\downarrow$  & Abs Rel $\downarrow$ & $\delta_1 \uparrow$ & mIoU $\uparrow$   \\ 
\midrule
\multirow{6}{*}{KITTI}      & Baseline                                                                  & $2.295$           & $0.056$              & $0.966$             & $0.615$           \\
                            & SW-Map~\cite{SW-Map}                                                                       & $2.275$           & $0.055$              & $0.967$             & $0.623$           \\
                            & SIG~\cite{SIG}                                                                       & $2.105$           & $0.053$              & $0.969$             & $0.650$           \\
                            & AG-MMD~\cite{AG-MMD}                                                                    & $2.096$           & $0.052$              & $0.973$             & $0.677$           \\
                            & CCAM~\cite{AffineMix}                                                                      & $2.080$           & $0.050$              & $0.976$             & $0.695$           \\ 
\cmidrule{2-6}
                            & \begin{tabular}[c]{@{}l@{}}\textbf{DSS}$^{*}$\\\end{tabular} & $\mathbf{1.984}$ & $\mathbf{0.048}$     & $\mathbf{0.979}$    & $\mathbf{0.731}$  \\ 
\midrule
\multirow{6}{*}{NYUv2}      & Baseline                                                                  & $0.323$           & $0.089$              & $0.934$             & $0.519$           \\
                            & SW-Map~\cite{SW-Map}                                                                       & $0.314$           & $0.087$              & $0.933$             & $0.522$           \\
                            & SIG~\cite{SIG}                                                                       & $0.309$           & $0.086$              & $0.937$             & $0.555$           \\
                            & AG-MMD~\cite{AG-MMD}                                                                    & $0.302$           & $0.085$              & $0.939$             & $0.571$           \\
                            & CCAM~\cite{AffineMix}                                                                      & $0.300$           & $0.083$              & $0.943$             & $0.589$           \\ 
\cmidrule{2-6}
                            & \begin{tabular}[c]{@{}l@{}}\textbf{DSS}$^{*}$\\\end{tabular} & $\mathbf{0.289}$  & $\mathbf{0.080}$     & $\mathbf{0.948}$    & $\mathbf{0.620}$  \\
\bottomrule
\end{tabular}}
\end{table}

\subsubsection{NearFarMix Augmentation}
\begin{figure*}[t]
    \centering
    \captionsetup{justification=centering}
    \includegraphics[scale=0.48]{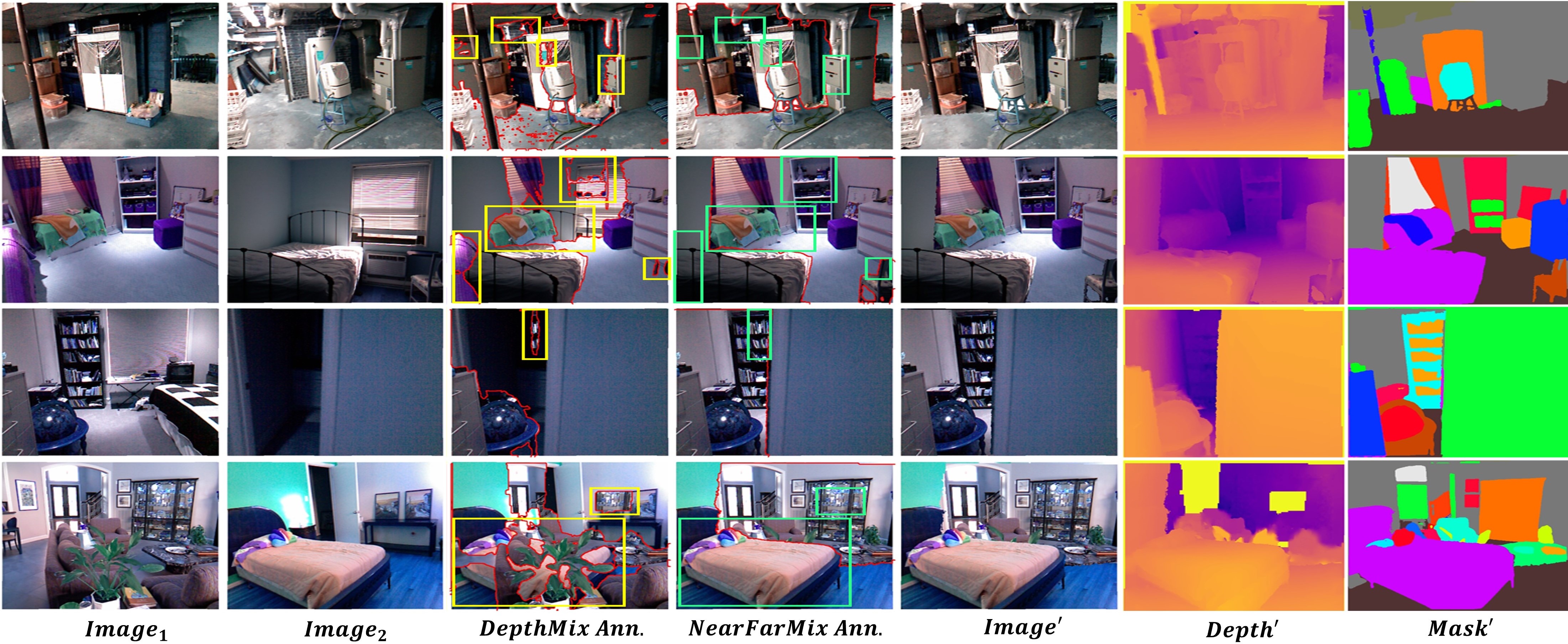}
    \caption{Visual contrasts between NearFarMix and DepthMix on NYUv2, with box areas showing of lost context or object integrity.}
    \label{fig:nearfarmix-nyuv2-quality}
\end{figure*}

\begin{figure*}[t]
    \centering
    \captionsetup{justification=centering}
    \includegraphics[scale=0.48]{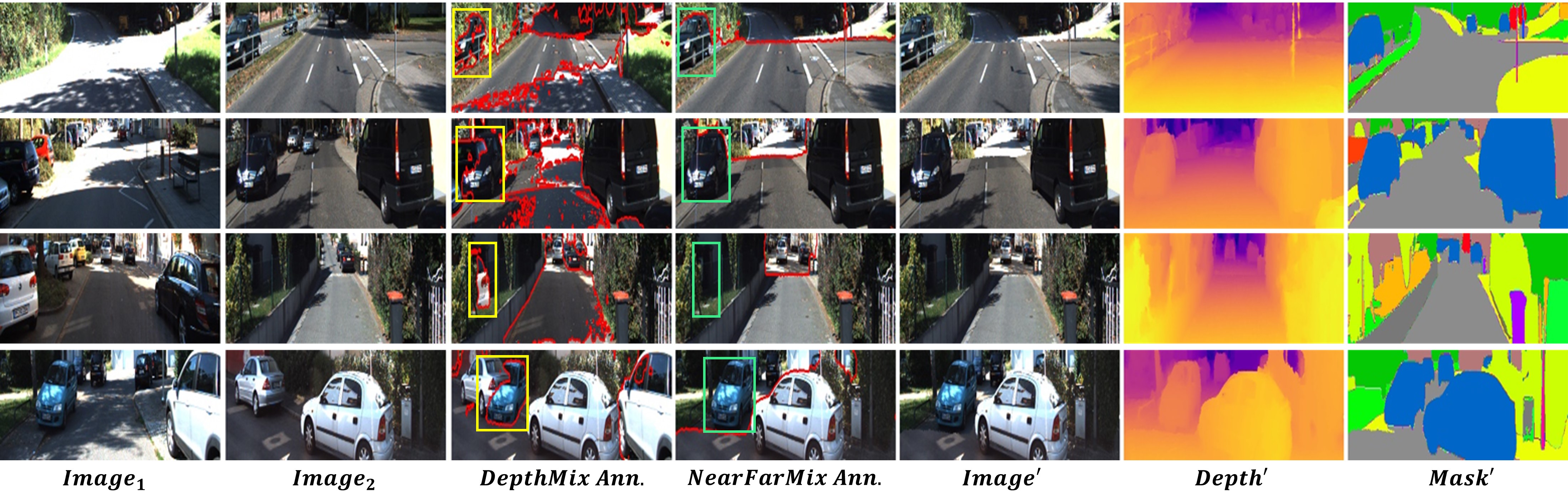}
    \caption{Visual contrasts between NearFarMix and DepthMix on KITTI, with box areas showing of lost context or object integrity.}
    \label{fig:nearfarmix-kitti-quality}
\end{figure*}

Table~\ref{tab:nearfarmix-ablation} contrasts the proposed NearFarMix augmentation with prevailing methods, utilizing the DSS module as a standard. The 'baseline' denotes our method without NearFarMix. Existing augmentations, enhancing semantics or depth, compromise either object integrity (e.g., CutMix~\cite{CutMix}, ClassMix~\cite{ClassMix}, AffineMix~\cite{AffineMix}, DepthMix~\cite{DepthMix}) or create semantic discontinuity (CutDepth~\cite{Cut-Depth}, V-CutDepth~\cite{GLPDepth}). NearFarMix balances depth and semantics, preserves integrity, and controls sampling diversity.  Visual evidence of NearFarMix's superiority over DepthMix~\cite{DepthMix} augmentation is provided in Fig. \ref{fig:nearfarmix-kitti-quality} and Fig. \ref{fig:nearfarmix-nyuv2-quality}, where the rectangle regions emphasize the method's immunity to object integrity and context loss.

\section{Conclusion}
\label{sec:conclusion}
This paper brings forth an efficacious semi-supervised method designed to address the challenge of semantic information scarcity across datasets while establishing a dataset-agnostic, generic architecture. Furthermore, it introduces a novel transformer, adept at harnessing the symbiotic potential between depth and semantics, unlocking a richer, integrated scene understanding through comprehensive local-global information sharing. Also presented is a novel augmentation, NearFarMix, purpose-built to enhance both depth and semantics tasks, while upholding structural integrity and avoiding semantic inconsistency. Robust experimental validation underscores the superiority of our method over existing techniques, as clearly demonstrated by significant performance improvements on both NYUv2 and KITTI datasets.

{\small
\bibliographystyle{ieee_fullname}
\bibliography{egbib}
}

\end{document}


\title{Supplementary Material}

\author{Md Awsafur Rahman, Shaikh Anowarul Fattah\\
Dept. of EEE, BUET, Bangladesh\\
{\tt\small \{1706066,fattah\}@eee.buet.ac.bd}
}
\maketitle

\maketitle

\section{NearFarMix Augmentation}

\begin{figure*}
    \centering
    \captionsetup{justification=raggedright}
    \includegraphics[scale=0.0820]{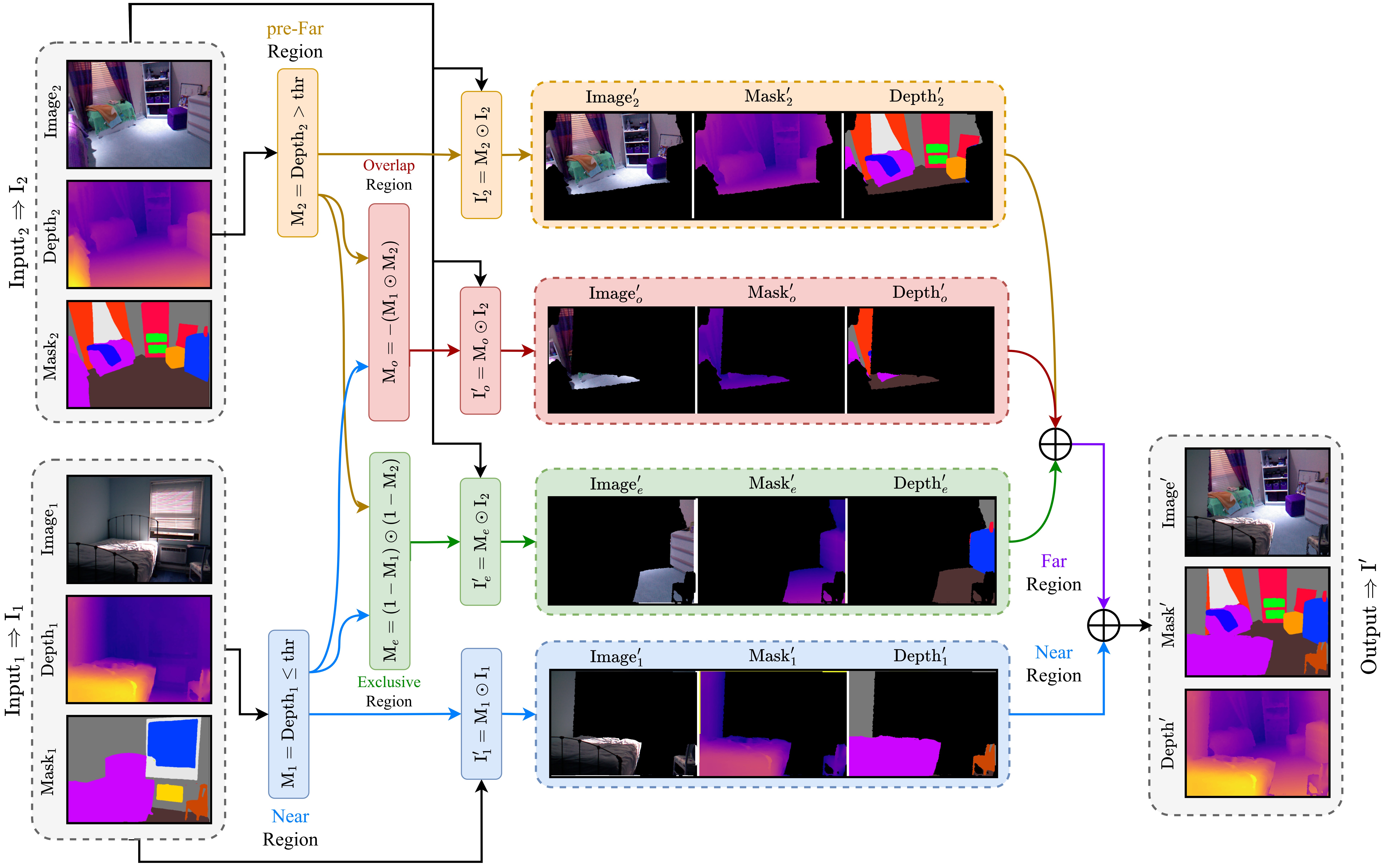}
    \caption{Proposed NearFarMix augmentation. The depth map undergoes thresholding to generate \textbf{Near} and \textbf{pre-Far} regions. Then, these regions are manipulated to produce \textbf{Overlap} and \textbf{Exclusive} regions. The \textbf{Far} region is then generated by combining the \textbf{pre-Far}, \textbf{Overlap}, and \textbf{Exclusive} regions, with the \textbf{Exclusive} region being subtracted and the remaining regions added. Finally, the \textbf{Near} and \textbf{Far} regions are combined to generate the augmented image.}
    \label{fig:nearfarmix}
\end{figure*}

\Cref{fig:nearfarmix} elegantly illustrates the operation of the proposed NearFarMix augmentation, delineating the four regions—near, pre-far, overlap, and exclusive—that contribute to the final augmented output. Despite the amalgamation of four regions, the pre-far, overlap, and exclusive regions primarily generate far regions, with overlap regions consequently subtracted. Importantly, the augmentation is designed for batch-wise application to enhance execution speed. The step-by-step implementation is comprehensively depicted in \Cref{algo:nearfarmix}.

\begin{figure*}
    \centering
    \captionsetup{justification=raggedright}
    \includegraphics[scale=0.480]{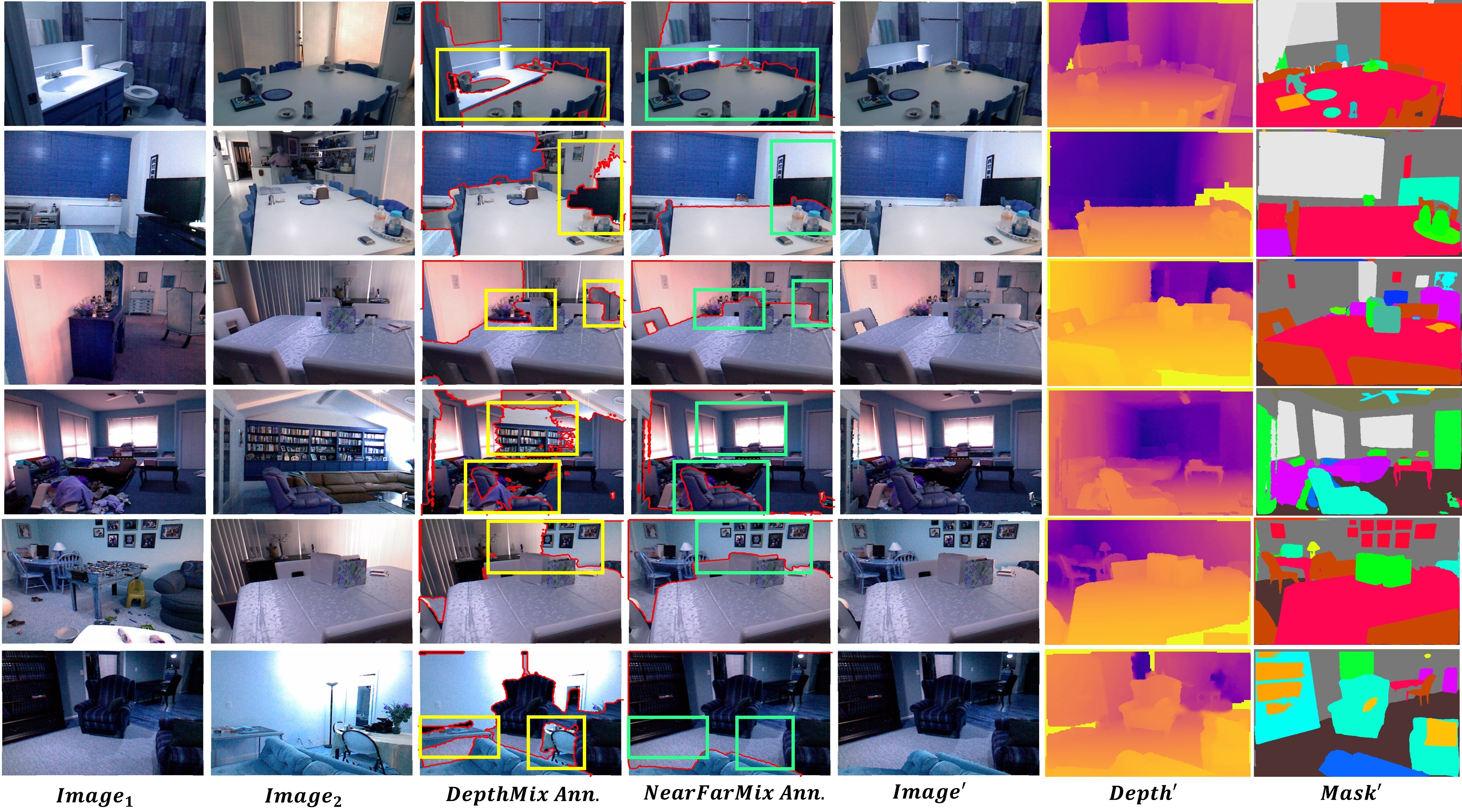}
    \caption{Additional qualitative comparisons between proposed NearFarMix and DepthMix augmentation on NYUv2 dataset.}
    \label{fig:suppl-nearfarmix-nyuv2-quality}
\end{figure*}

\begin{figure*}
    \centering
    \captionsetup{justification=raggedright}
    \includegraphics[scale=0.480]{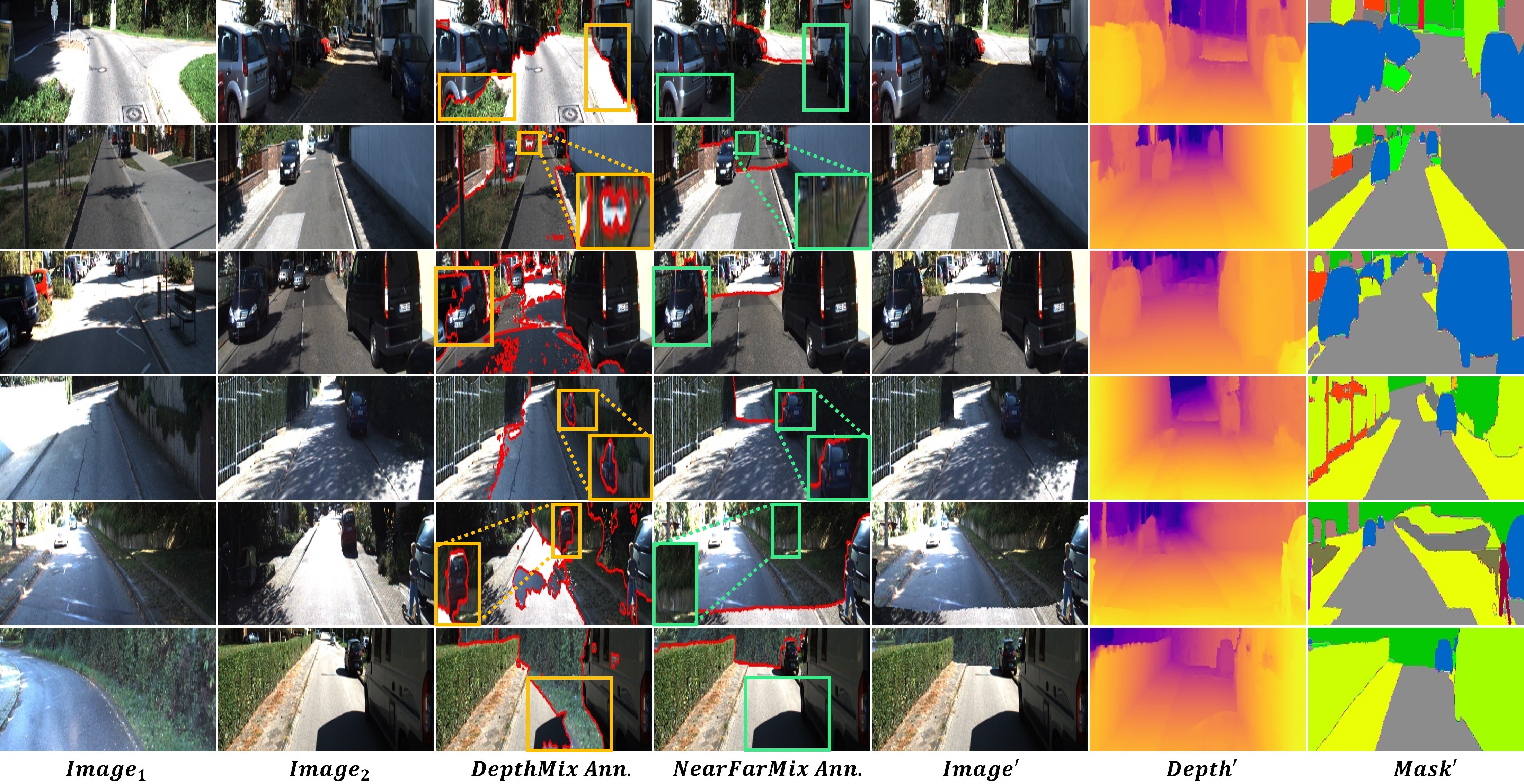}
    \caption{Additional qualitative comparisons between proposed NearFarMix and DepthMix augmentation on KITTI dataset.}
    \label{fig:suppl-nearfarmix-kitti-quality}
\end{figure*}

Additionally, \cref{fig:suppl-nearfarmix-kitti-quality} and \cref{fig:suppl-nearfarmix-nyuv2-quality} provide additional examples between the proposed NearFarMix and DepthMix~\cite{DepthMix}.

\begin{algorithm}
\caption{Batchwise Window Partition}\label{algo:window}
\begin{algorithmic}
\State $x \gets Input$ \Comment{Features}
\State $h_w, w_w \gets window\_size$ \Comment{Size of each window}
\State
\State \textbf{\# Partition features into local dense windows}
\State $B, H, W, C = \text{shape}(x)$
\State $x = \text{reshape}(x, shape=[B, \frac{H}{h_w}, h_w, \frac{W}{w_w}, w_w, C])$
\State $x = \text{transpose}(x, permute\_axis=[0, 1, 3, 2, 4, 5])$
\State $x = \text{reshape}(x, shape=[B \times \frac{H \times W}{h_w \times w_w}, h_w, w_w, C])$
\end{algorithmic}
\end{algorithm}

\begin{algorithm}
\caption{Batchwise Grid Partition}\label{algo:grid}
\begin{algorithmic}
\State $x \gets Input$ \Comment{Features}
\State $h_g, w_g \gets grid\_size$ \Comment{Size of each grid}
\State
\State \textbf{\# Partition features into global sparse grids} 
\State $B, H, W, C = \text{shape}(x)$ 
\State $x = \text{reshape}(x, shape=[B, h_g, \frac{H}{h_g}, w_g, \frac{W}{w_g}, C])$ 
\State $x = \text{transpose}(x, permute\_axis=[0, 1, 3, 2, 4, 5])$ 
\State $x = \text{reshape}(x, shape=[B, h_g \times w_g, \frac{H \times W}{h_g \times w_g}, C])$ 
\State $x = \text{transpose}(x, permute\_axis=[0, 2, 1, 3])$ 
\State $x = \text{reshape}(x, shape=[B \times \frac{H \times W}{h_g \times w_g}, h_g, w_g, C])$ 
\end{algorithmic}
\end{algorithm}

\begin{algorithm}
\caption{Batchwise NearFarMix Augmentation}\label{algo:nearfarmix}
\begin{algorithmic}

\State $I_1 \gets Images$  \Comment{Input images}
\State $D_1 \gets Depths$  \Comment{Input depths}
\State $S_1 \gets Semantics$  \Comment{Input semantics}
\State $\mathcal{U} \gets random\_uniform$  \Comment{Uniform distribution}
\State $D_{min} \gets 20 (\text{KITTI}) \quad\text{or} \quad 1.5 (\text{NYUv2})$  \Comment{Min depth}
\State $D_{max} \gets 60 (\text{KITTI}) \quad\text{or} \quad 6.5 (\text{NYUv2})$  \Comment{Max depth}
\State

\State \textbf{\# Roll batch for fast shuffling} 
\State $I_2 = \operatorname{roll}(I_1, shift=1, axis=0)$  \Comment{Roll Images}
\State $D_2 = \operatorname{roll}(D_1, shift=1, axis=0)$  \Comment{Roll Depths}
\State $S_2 = \operatorname{roll}(S_1, shift=1, axis=0)$  \Comment{Roll Semantics}
\State

\State \textbf{\# Depth threshold range for batch} 
\State  $d_{min} = \operatorname{max}(\operatorname{min}(D_1, axis=(1, 2, 3)))$  \Comment{Min depth}
\State  $d_{max} = \operatorname{min}(\operatorname{max}(D_1, axis=(1, 2, 3)))$  \Comment{Max depth}
\State

\State \textbf{\# Threshold for Near-Far region} 
\State $B, H, W, C = shape(I_1)$  
\State $thr_{min} = max(D_{min}, d_{min})$ \Comment{Clip min depth}
\State $thr_{max} = min(D_{max}, d_{max}) $ \Comment{Clip max depth}
\State $thrs = \mathcal{U}(shape = [B, 1, 1, 1],$  \Comment{Random thresholds}
\Statex \hspace{1.5cm} $min=thr_{min},$
\Statex \hspace{1.5cm} $max=thr_{max})$
\State

\State \textbf{\# Compute binary masks of regions for blending} 
\State $M_1 = D_1<=thrs$  \Comment{Broadcasted Near region mask}
\State $M_2 = D_2>thrs$  \Comment{Broadcasted pre-Far region mask}
\State $M_o = M_1 \odot M_2$  \Comment{Overlap region mask}
\State $M_e = (1 - M_1) \odot (1 - M_2)$  \Comment{Exclusive region mask}
\State

\State \textbf{\# Perform blending of regions}
\State $I' = (I_1 \odot M_1)_{near} + ((I_2 \odot M_2)  + (I_2 \odot M_e) - (I_2 \odot M_o))_{far}$  \Comment{Augmented image}
\State $D' = (D_1 \odot M_1)_{near} + ((D_2 \odot M_2) + (D_2 \odot M_e) - (D_2 \odot M_o))_{far}$  \Comment{Augmented depth}
\State $S' = (S_1 \odot M_1)_{near} + ((S_2 \odot M_2) + (S_2 \odot M_e) - (S_2 \odot M_o))_{far}$  \Comment{Augmented semantics}
\end{algorithmic}
\end{algorithm}

\begin{algorithm}
\caption{Local-Global Cross-Attention Transformer (LG-CAT)}\label{algo:lg-cat}
\begin{algorithmic}
\State $F_x^{Q}, F_y^{KV} \gets inputs$ \Comment{Input features}
\State $x \gets F_x^{Q}$ \Comment{Query features of depth/semantics}
\State $y \gets F_y^{KV}$ \Comment{Key-value features of semantics/depth}
\State $i \gets 0$ \Comment{Initialize counter}
\State

\While{$i \neq 2$}
\State \textbf{\# Block Cross Attention}
\State $x_1 = \text{layer\_norm}(x)$
\State $y_1 = \text{layer\_norm}(y)$
\State $x_{1,q} = \text{FFN}(\text{window\_partition}(x_1))$ \Comment{Query gen.}
\State $y_{1,k}, y_{1,v} = \text{FFN}(\text{window\_partition}(y_1))$ \Comment{KeyValue}
\State $y_2 = \text{CA}(x_{1,q}, y_{1,k}, y_{1,v})$ \Comment{Apply cross-attention}
\State $y_2 = y_1 + \text{window\_reverse}(\text{FFN}(y_2))$ \Comment{Residual}
\State

\State \textbf{\# Grid Cross Attention}
\State $y_2 = \text{layer\_norm}(y_2)$
\State $x_{2,q} = \text{FFN}(\text{grid\_partition}(x_1))$ \Comment{Query}
\State $y_{2,k}, y_{2,v} = \text{FFN}(\text{grid\_partition}(y_2))$ \Comment{KeyValue}
\State $y_3 = \text{CA}(x_{2,q}, y_{2,k}, y_{2,v})$ \Comment{Apply cross-attention}
\State $y_3 = y_2 + \text{grid\_reverse}(\text{FFN}(y_3))$ \Comment{Residual}
\State $y = y_3$ \Comment{Reset variable for loop}
\State $i = i+1$ \Comment{Increment counter}

\EndWhile
\State

\State \textbf{\# FusedMBConv - Channel Attention}
\State $\hat{y} = \text{DWConv3x3}(y)$ \Comment{Depthwise convolution}
\State $\hat{y} = \text{GELU}(\hat{y})$ \Comment{Apply activation}
\State $\hat{y} = \text{SE}(\hat{y})$ \Comment{Squeeze-Excitation}
\State $\hat{y} = \text{Conv1x1}(\hat{y})$ \Comment{Convolution}
\State $\hat{y} = y + \hat{y}$ \Comment{Residual}
\State

\State $output \gets \hat{y}$

\end{algorithmic}
\end{algorithm}

\vspace{10em}
\section{Symbiotic Transformer}

\textbf{Transformers:} \Cref{eq:sym-t} presents the detailed mathematical expression for Symbiotic Transformer, which symbiotically enhances both depth and semantics via local-global cross-attention. In the equation, $\mathbf{F_x^{Q}}$ represents query features of $\mathbf{x}$, $\mathbf{F_y^{KV}}$ denotes key-value features of $\mathbf{y}$, and $\mathbf{F_y}$ signifies the output features contextualized by $\mathbf{x}$. Specifically, for SGT, $\mathbf{x}=\mathbf{s}$ and $\mathbf{y}=\mathbf{d}$, while for DGT, $\mathbf{x}=\mathbf{d}$ and $\mathbf{y}=\mathbf{s}$. Moreover, $\operatorname{DGT} = \operatorname{LG-CAT}_{{}_{DG}}$ and $\operatorname{SGT} = \operatorname{LG-CAT}_{{}_{SG}}$ correspond to depth and semantics-guided local-global cross-attention transformers.

\textbf{Cross Attentions:}  Under the hood, DGT and SGT employ semantics-guided cross attention (SG-CA) and depth-guided cross attention (DG-CA), respectively to contextualize features. SG-CA and DG-CA is mathematically elaborated in \cref{eq:sg-ca} and \cref{eq:dg-ca}.  In these equations, $\mathbf{W}$ represents the weight of the $\operatorname{FFN}$ layer, 
$\mathbf{Q_x}$ represents query features, $\mathbf{K_y}$ and $\mathbf{V_y}$ represent key and value features, $\mathcal{S}mx$ denotes softmax, and $\mathbf{d}$ is the query/key dimension. $\mathbf{B}$ represents relative positional bias, sampled similar to~\cite{GCViT}. The Local-Global Cross-Attention Transformer (LG-CAT), employed by both SGT and DGT, can be implemented using \Cref{algo:lg-cat}.

\textbf{Partition Operation: }Further, the algorithms for $\operatorname{Window Partition}$ and $\operatorname{Grid Partition}$ operations which are used in Block and Grid attention also slightly different from the methods of Max-ViT~\cite{MaxViT}, are detailed in \cref{algo:window} and \cref{algo:grid}, respectively. It is noteworthy that Max-ViT implemented these operations using einops~\cite{einops}.

\begin{equation}
\vspace*{10cm}
\label{eq:sym-t}
\scalemath{0.90}{
\operatorname{Sym-T}(\mathbf{F'_d, F'_s}) = \begin{bmatrix}
&\mathbf{F_d^{Q}} = \mathbf{F_d^{KV}} = \mathbf{F'_d}\\
&\mathbf{F_s^{Q}} = \mathbf{F_s^{KV}} = \mathbf{F'_s}\\
\mathbf{F_s} &= \operatorname{LG-CAT}_{{}_{DG}}(\mathbf{F_d^{Q}}, \mathbf{F_s^{KV}}) \\
\mathbf{F_d} &= \operatorname{LG-CAT}_{{}_{SG}}(\mathbf{F_s^{Q}}, \mathbf{F_d^{KV}}) \\
\end{bmatrix}}
\end{equation}

\begin{equation}
\scalemath{0.78}{
\begin{aligned}
\label{eq:sg-ca}
\operatorname{SG-CA} &= \operatorname{CA}\left(\mathbf{Q_s}, \mathbf{K_d}, \mathbf{V_d}\right)\\ 
&= \mathcal{S}mx\left(\frac{\mathbf{Q_s} {\mathbf{K_d}}^T}{\sqrt{\mathbf{d}}}+\mathbf{B}\right)\mathbf{V_d}\\
&= \mathcal{S}mx\left(\frac{(\mathbf{F_s^{Q} \cdot W_Q^{s}}) ({\mathbf{F_d^{KV} \cdot W_K^{d}}})^T}{\sqrt{\mathbf{d}}}+\mathbf{B}\right)(\mathbf{F_d^{KV} \cdot W_V^{d}})\\
\end{aligned}}
\end{equation}
\vspace*{-0.2cm}

\begin{equation}
\scalemath{0.78}{
\begin{aligned}
\label{eq:dg-ca}
\operatorname{DG-CA} &= \operatorname{CA}\left(\mathbf{Q_d}, \mathbf{K_s}, \mathbf{V_s}\right)\\
&= \mathcal{S}mx\left(\frac{\mathbf{Q_d} {\mathbf{K_s}}^T}{\sqrt{\mathbf{d}}}+\mathbf{B}\right)\mathbf{V_s}\\
&= \mathcal{S}mx\left(\frac{(\mathbf{F_d^{Q} \cdot W_Q^{d}}) ({\mathbf{F_s^{KV} \cdot W_K^{s}}})^T}{\sqrt{\mathbf{d}}}+\mathbf{B}\right)(\mathbf{F_s^{KV} \cdot W_V^{s}})\\
\end{aligned}}
\end{equation}

\section{Architecture Details}
The architectural specifications, encompassing input, output, layer name, and layer details, are succinctly laid out in \Cref{tab:arch-config}. Here, $E$ and $D$ represent the input/output of the encoder/decoder, while $ST$ and $N$ denote to the Symbiotic Transformer and the Neck.

\begin{table*}
\centering
\captionsetup{justification=raggedright,singlelinecheck=false}
\caption{Architectural Specifications of proposed method where $h$, $w$ signify attention heads and window size; $Conv$ denotes 2D convolution with $k$, $s$, $c$ as kernel size, stride size, and output channels; $act$ and $norm$ represent activation and normalization types; $sc$ indicates upscale size.}
\label{tab:arch-config}
\scalebox{0.90}{
\begin{tabular}{c|c|c|c|c} 
\toprule
\multicolumn{5}{c}{Input Size: $H \times W \times$ 3}                                                                                                                                                                                                                                                                                                                                                                                                                                                                                                                     \\ 
\toprule
Layer Name                                                                      & Input                             & Output                              & Output Size                                                                                                                                          & Architecture                                                                                                                                                                                                                                           \\ 
\cmidrule{1-1}\cline{2-4}\cmidrule{5-5}
\multirow{2}{*}{Stem}                                                           & \multirow{2}{*}{$Image$}          & \multirow{2}{*}{$E_0$}              & \multirow{2}{*}{$\frac{H}{4} \times \frac{W}{2} \times 128$}                                                                                         & Conv(c=128, k=3, s=2)                                                                                                                                                                                                                                  \\ 
\cline{5-5}
                                                                                &                                   &                                     &                                                                                                                                                      & Conv(c=128, k=3, s=1)                                                                                                                                                                                                                                  \\ 
\hline
\begin{tabular}[c]{@{}c@{}}Encoder\\Stage 1\end{tabular}                        & $E_0$                             & $E_1$                               & $\frac{H}{8} \times \frac{W}{4} \times 128$                                                                                                          & $\begin{bmatrix}\text{Max-ViT-Block(h=4, w=7)}\end{bmatrix} \times 2$                                                                                                                                                                                   \\ 
\hline
\begin{tabular}[c]{@{}c@{}}Encoder\\Stage 2\end{tabular}                        & $E_1$                             & $E_2$                               & $\frac{H}{16} \times \frac{W}{8} \times 256$                                                                                                         & $\begin{bmatrix}\text{Max-ViT-Block(h=8, w=7)}\end{bmatrix} \times 6$                                                                                                                                                                                   \\ 
\hline
\begin{tabular}[c]{@{}c@{}}Encoder\\Stage 3\end{tabular}                        & $E_2$                             & $E_3$                               & $\frac{H}{32} \times \frac{W}{16} \times 512$                                                                                                        & $\begin{bmatrix}\text{Max-ViT-Block(h=16, w=7)}\end{bmatrix} \times 14$                                                                                                                                                                                 \\ 
\hline
\begin{tabular}[c]{@{}c@{}}Encoder\\Stage 4\end{tabular}                        & $E_3$                             & $E_4 $/ $D_4$                       & $\frac{H}{32} \times \frac{W}{32} \times 1024$                                                                                                       & $\begin{bmatrix}\text{Max-ViT-Block(h=16, w=7)}\end{bmatrix} \times 2$                                                                                                                                                                                  \\ 
\cline{1-3}\cmidrule{4-4}\cline{5-5}
\multirow{3}{*}{\begin{tabular}[c]{@{}c@{}}Decoder\\Stage 3\end{tabular}}       & \multirow{3}{*}{$(D_4, E_3)$}     & \multirow{3}{*}{$D_3$}              & \multirow{3}{*}{$\frac{H}{32} \times \frac{W}{32} \times 512$}                                                                                       & Upsample(sc=2)                                                                                                                                                                                                                                         \\ 
\cline{5-5}
                                                                                &                                   &                                     &                                                                                                                                                      & Concat($[E_3, D_4]$, axis=-1)                                                                                                                                                                                                                          \\ 
\cline{5-5}
                                                                                &                                   &                                     &                                                                                                                                                      & Conv(c=512, k=3, s=1, norm='layer', act='gelu')                                                                                                                                                                                                        \\ 
\cline{1-4}\cmidrule{5-5}
\multirow{3}{*}{\begin{tabular}[c]{@{}c@{}}Decoder\\Stage 3\end{tabular}}       & \multirow{3}{*}{$(D_3, E_2)$}     & \multirow{3}{*}{$D_2$}              & \multirow{3}{*}{$\frac{H}{16} \times \frac{W}{16} \times 256$}                                                                                       & Upsample(sc=2)                                                                                                                                                                                                                                         \\ 
\cline{5-5}
                                                                                &                                   &                                     &                                                                                                                                                      & Concat($[E_2, D_3]$, axis=-1)                                                                                                                                                                                                                          \\ 
\cline{5-5}
                                                                                &                                   &                                     &                                                                                                                                                      & Conv(c=256, k=3, s=1, norm='layer', act='gelu')                                                                                                                                                                                                        \\ 
\hline
\multirow{3}{*}{\begin{tabular}[c]{@{}c@{}}Decoder\\Stage 1\end{tabular}}       & \multirow{3}{*}{$(D_2, E_1)$}     & \multirow{3}{*}{$D_1$}              & \multirow{3}{*}{$\frac{H}{4} \times \frac{W}{4} \times 128$}                                                                                         & Upsample(sc=2)                                                                                                                                                                                                                                         \\ 
\cline{5-5}
                                                                                &                                   &                                     &                                                                                                                                                      & Concat($[E_1, D_2]$, axis=-1)                                                                                                                                                                                                                          \\ 
\cline{5-5}
                                                                                &                                   &                                     &                                                                                                                                                      & Conv(c=128, k=3, s=1, norm='layer', act='gelu')                                                                                                                                                                                                        \\ 
\hline
Neck                                                                            & $D_1$                             & $(N_d, N_s)$                        & \begin{tabular}[c]{@{}c@{}}$(\frac{H}{4} \times \frac{W}{4} \times 150,$\\$\frac{H}{4} \times \frac{W}{4} \times 150)$\end{tabular}                  & \begin{tabular}[c]{@{}c@{}}$(\begin{bmatrix}\text{Conv(c=150, k=3, s=1, norm='layer', act='gelu')}\end{bmatrix} \times 2, $\\$\begin{bmatrix}\text{Conv(c=150, k=3, s=1, norm='layer', act='gelu')}\end{bmatrix} \times 2)$\end{tabular}               \\ 
\cmidrule{1-1}\cline{2-3}\cmidrule{4-5}
\multirow{2}{*}{\begin{tabular}[c]{@{}c@{}}Symbiotic\\Transformer\end{tabular}} & \multirow{2}{*}{$(N_{d}, N_{s})$} & \multirow{2}{*}{$(ST_{d}, ST_{s})$} & \multirow{2}{*}{\begin{tabular}[c]{@{}c@{}}$(\frac{H}{4} \times \frac{W}{4} \times 150,$\\$\frac{H}{4} \times \frac{W}{4} \times 150)$\end{tabular}} & $\begin{bmatrix} \text{Block-Cross-Attention(h=4, w=7)} \\ \text{Grid-Cross-Attention(h=4, w=7)} \end{bmatrix} \times 2$                                                                                                                               \\ 
\cmidrule{5-5}
                                                                                &                                   &                                     &                                                                                                                                                      & $\begin{bmatrix}\text{FusedMBConv}\end{bmatrix} \times 1$                                                                                                                                                                                              \\ 
\cmidrule{1-1}\cline{2-3}\cmidrule{4-5}
Head                                                                            & $(ST_d, ST_s)$                    & $(Depth, Semantics)$                & \begin{tabular}[c]{@{}c@{}}$(H \times W \times 1,$\\$H \times W \times 150)$\end{tabular}                                                            & \begin{tabular}[c]{@{}c@{}}$(\begin{bmatrix}\text{Conv(k=3, s=1, act='sigmoid')} \\ \text{Upsample(sc=4)}\end{bmatrix} \times 2,$\\$\begin{bmatrix}\text{Conv(k=3, s=1, act='softmax')} \\ \text{Upsample(sc=4)}\end{bmatrix} \times 2)$\end{tabular}  \\ 
\cmidrule{1-1}\cline{2-3}\cmidrule{4-5}
\multicolumn{5}{c}{($Depth$: $H \times W \times 1$, $Semantics$: $H \times W \times 150$)}                                                                                                                                                                                                                                                                                                                                                                                                                                                                                \\
\bottomrule
\end{tabular}}
\end{table*}

\newpage

{\small
\bibliographystyle{ieee_fullname}
\bibliography{egbib}
}